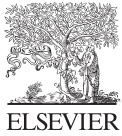
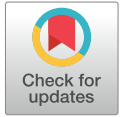
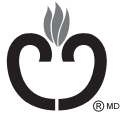

## Original Article

# Machine Learning Methods for Identifying Atrial Fibrillation Cases and Their Predictors in Patients With Hypertrophic Cardiomyopathy: The HCM-AF-Risk Model

Moumita Bhattacharya, PhD,[a,‡] Dai-Yin Lu, MD,[b,c,d,e,‡] Ioannis Ventoulis, MD,[b]
Gabriela V. Greenland, MD,[b,e] Hulya Yalcin, MD,[b] Yufan Guan, MD,[b] Joseph E. Marine, MD,[b]
Jeffrey E. Olgin, MD,[e] Stefan L. Zimmerman, MD,[f] Theodore P. Abraham, MD,[b,e]
M. Roselle Abraham, MD,[b,e,§] and Hagit Shatkay, PhD[a,§]

[a] Computational Biomedicine and Machine Learning Lab, Department of Computer and Information Sciences, University of Delaware, Newark, Delaware, USA
[b] Hypertrophic Cardiomyopathy Center of Excellence, Johns Hopkins University, Baltimore, Maryland, USA
[c] Division of General Medicine, Taipei Veterans General Hospital, Taipei, Taiwan
[d] Institute of Public Health, National Yang-Ming University, Taipei, Taiwan
[e] Hypertrophic Cardiomyopathy Center of Excellence, Division of Cardiology, University of California San Francisco, San Francisco, California, USA
[f] Department of Radiology, Johns Hopkins University, Baltimore, Maryland, USA

## ABSTRACT

**Background:** Hypertrophic cardiomyopathy (HCM) patients have a high incidence of atrial fibrillation (AF) and increased stroke risk, even with low CHA$_2$DS$_2$-VASc (congestive heart failure, hypertension, age diabetes, previous stroke/transient ischemic attack) scores. Hence, there is a need to understand the pathophysiology of AF/stroke in HCM. In

## RÉSUMÉ

**Introduction :** Les patients atteints d'une cardiomyopathie hypertrophique (CMH) présentent une forte incidence de fibrillation auriculaire (FA) et un risque accru d'accident vasculaire cérébral (AVC), malgré des scores CHA$_2$DS$_2$-VASc (*congestive heart failure, hypertension, age diabetes, previous stroke/transient ischemic attack,*

Hypertrophic cardiomyopathy (HCM) is characterized by myocyte hypertrophy, myocyte disarray, and interstitial/replacement fibrosis, and it is associated with a high risk for atrial and ventricular arrhythmias. A large proportion (∼25%-30%) of HCM patients develop atrial fibrillation (AF) during their lifetime.[1] Notably, HCM patients with AF are at high risk for stroke even in the setting of low Congestive Heart Failure, Hypertension, Age, Diabetes, Previous Stroke/Transient Ischemic Attack, Vascular Disease (Prior

Myocardial Infarction, Peripheral Artery Disease, Aortic Plaque), Sex (CHA$_2$DS$_2$-VASc) scores.[1-3] Furthermore, stroke can be the first manifestation of AF in HCM. Hence, there is a need to identify risk factors for AF and stroke in HCM.[4] In order to predict risk, an improved understanding of the pathophysiology of AF in HCM is needed. The first step in AF risk prediction is identification of clinical and imaging features associated with AF in HCM patients.

Electronic health records provide a wealth of biological and physiological data that can be used for computerized phenotyping of patients, using machine learning. In contrast to rule-based methods (traditional models),[5] machine learning allows for thorough scanning of clinical data along several dimensions and serves as a basis for classification algorithms that can stratify cases based on their respective likelihood to present with disease (AF in this case) equally. An advantage of machine learning–based algorithms is their robustness and ease of updating as additional clinical data become available.

In this retrospective study, we used electronic health record data of HCM patients who underwent deep clinical phenotyping by multi-modality imaging, to identify clinical and imaging features associated with higher/lower risk of AF, using machine learning. However, since the number of AF cases is this retrospective study, we develop and apply a data-driven, machine learning–based method to identify AF cases, and clinical/imaging features associated with AF, using electronic health record data.

**Methods:** HCM patients with documented paroxysmal/persistent/ permanent AF (n = 191) were considered AF cases, and the remaining patients in sinus rhythm (n = 640) were tagged as No-AF. We evaluated 93 clinical variables; the most informative variables useful for distinguishing AF from No-AF cases were selected based on the 2-sample *t* test and the information gain criterion.

**Results:** We identified 18 highly informative variables that are positively (n = 11) and negatively (n = 7) correlated with AF in HCM. Next, patient records were represented via these 18 variables. Data imbalance resulting from the relatively low number of AF cases was addressed via a combination of oversampling and undersampling strategies. We trained and tested multiple classifiers under this sampling approach, showing effective classification. Specifically, an ensemble of logistic regression and naïve Bayes classifiers, trained based on the 18 variables and corrected for data imbalance, proved most effective for separating AF from No-AF cases (sensitivity = 0.74, specificity = 0.70, C-index = 0.80).

**Conclusions:** Our model (HCM-AF-Risk Model) is the first machine learning–based method for identification of relevant cases in HCM. This model demonstrates good performance, addresses data imbalance, and suggests that AF is associated with a more severe cardiac HCM phenotype.

c'est-à-dire : insuffisance cardiaque congestive, hypertension, âge, diabète, AVC ou accident ischémique transitoire antérieur) faibles. Par conséquent, il est nécessaire de comprendre la physiopathologie de la FA et de l'AVC en présence d'une CMH. Dans la présente étude rétrospective, nous avons élaboré et appliqué une méthode d'apprentissage automatique dirigée sur les données pour déterminer les cas de FA, et les caractéristiques cliniques/d'imagerie associées à la FA, à l'aide des données des dossiers de santé électroniques.

**Méthodes :** Nous avons considéré les patients atteints d'une CMH qui ont une FA paroxystique/persistante/permanente documentée (n = 191) comme les cas de FA, et avons étiqueté les autres patients en rythme sinusal (n = 640) comme des cas sans FA. Nous avons évalué 93 variables cliniques; nous avons sélectionné les variables les plus informatives qui sont utiles pour distinguer les cas de FA des cas sans FA en fonction du test t pour deux échantillons et du critère de gain d'information.

**Résultats :** Nous avons relevé 18 variables hautement informatives qui ont une corrélation positive (n = 11) et une corrélation négative (n = 7) avec la FA en présence d'une CMH. Ensuite, nous avons représenté les dossiers des patients au moyen de ces 18 variables. Nous avons remédié au déséquilibre des données, qui résulte du nombre relativement faible de cas de FA, grâce à une combinaison de stratégies de suréchantillonnage et de sous-échantillonnage. Nous avons formé et testé de nombreux classificateurs selon cette approche d'échantillonnage, qui montre une classification efficace. Particulièrement, un ensemble de régression logistique et de classificateurs bayésiens naïfs formés en fonction des 18 variables et corrigés en fonction du déséquilibre des données s'est révélé le plus efficace pour séparer les cas de FA des cas sans FA (sensibilité = 0,74, spécificité = 0,70, indice C = 0,80).

**Conclusions :** Notre modèle (modèle de risque de CMH-FA) est la première méthode d'apprentissage automatique qui sert à déterminer les cas de FA en présence de CMH. Ce modèle permet de démontrer une bonne performance, de remédier au déséquilibre des données, et de croire que la FA est associée à un phénotype grave de CMH.

small when compared to the overall number of HCM cases, the data we are handling are inherently imbalanced. Thus, there is a need to address data imbalance as part of the model development. The model we introduce (HCM-AF-Risk Model) explicitly handles data imbalance. Here, each HCM patient is represented as a collection (vector) of interpretable clinical values. The learned classification decision itself is based on the probability that a given patient has AF or history of AF. Our method stands in contrast to most recently published work in machine learning within the clinical domain,[6-8] for which a complex model architecture based on artificial neural networks is used. The latter acts as a "black box" that assigns the categorization label for the patient, without the ability to track down the justification or explanation.

## Methods

### Clinical data and outcomes

**Patient population.** Our HCM Registry (Johns Hopkins Hospital [JHH]-HCM Registry) is approved by the institutional review boards of the JHH and the University of California San Francisco. Patients were enrolled in the JHH-HCM Registry during their first visit to the JHH

HCM-Center of Excellence[9] if they met the standard diagnostic criteria for HCM, namely, maximal left ventricle (LV) wall thickness ≥15 mm[10] in the absence of uncontrolled hypertension, valvular heart disease, and HCM phenocopies (amyloidosis, storage disorders).[11]

We performed a retrospective study of all HCM patients from the JHH-HCM Registry who were evaluated between January 1, 2003 and March 31, 2017. Clinical data including symptoms, comorbidities, medications, and history of arrhythmias were ascertained by the examining physician (M.R.A., T.P.A.) during the initial clinic visit, and during each follow-up visit. Rest and treadmill exercise stress echocardiography (ECHO) and cardiac magnetic resonance imaging (CMR) were performed at the first visit as part of patients' clinical evaluation. Patients who were asymptomatic or had stable symptoms were followed yearly; symptomatic patients were followed more frequently (every 1-3 months) until symptom management was achieved. During yearly follow-up visits, patients underwent treadmill exercise ECHO and 24-hour Holter monitoring or implantable cardioverter defibrillator (ICD) interrogation. Patients with a pacemaker/ ICD had remote device monitoring and device interrogation performed every 6 months, or more frequently if they were symptomatic or experienced ICD discharges. Patients (without implanted devices) who had palpitations without evidence of arrhythmias on Holter monitor or exercise-electrocardiogram (EKG) were provided event monitors to document cardiac rhythm during symptoms. A subset of HCM patients (n = 145) were referred for perfusion 13N-ammonia positron emission tomography (perfusion-$^{13}$NH3-PET) imaging after ruling out obstructive coronary artery disease by coronary angiography—these pateints had angina, ventricular arrhythmias, and/or exertional dyspnea despite optimal therapy.

**Atrial fibrillaton.** AF was diagnosed if AF/flutter of any duration was present. Paroxysmal AF (PAF) was defined as AF that terminated spontaneously or with intervention in ≤ 7 days of AF onset[12]; persistent AF was defined as AF that lasted >7 days and terminated spontaneously or with treatment; permanent AF was defined as AF that persisted despite treatment to restore sinus rhythm.[13]

AF was diagnosed by review of rest/stress 12-lead EKGs, event recorder data, Holter monitor data, and/or ICD inter-rogation; a detailed chart review was performed to confirm this criterion. Patients with PAF had confirmed termination of AF within the 7-day window, either by Holter monitor or EKG. Review of medical records, Holter monitor/event recorder studies, and ICD interrogations were performed in patients from the No-AF group to ensure they had no documented history of AF prior to their first clinic visit or during follow up.

**Cardiac imaging.** Transthoracic echocardiography was performed using a GE Vivid 7 or E-9 ultrasound machine and a multifrequency phased-array transducer. Left atrial diameter (anteroposterior) was measured in the parasternal long-axis view at the level of the aortic sinuses, perpendicular to the aortic root long axis, by using the leading-edge to leading-edge convention, just before mitral valve opening (LV end-systole).[14] Echocar-diographic images for 2-dimensional speckle tracking strain analysis were acquired prospectively at frame rates of 50-90 Hz. Longitudinal strain/strain rate was analyzed from the apical 2-, 3-, and 4-chamber views using EchoPAC 112.[15] CMR imaging was performed at the first clinic visit, using a 1.5T system, with administration of the contrast agent gadopentetate dimeglumine (0.2 mmol/kg).[16] LV mass and late gadolinium enhancement (LV-LGE) were quantified using QMass software (QMass 7.4; Medis, Leiden, The Netherlands). Cardiac $^{13}$NH$_3$-PET/CT imaging was performed using a GE Discovery VCT PET/CT system and a 1-day rest/stress protocol, as described previously.[17] Please refer to the Supplemental Appendix S1, Section A1 for detailed methods for rest/stress echocardiography, CMR, and PET/CT imaging.

**Cardiovascular events during follow-up.** All analyses were blinded to AF outcome. Cardiovascular adverse events, including AF, stroke, heart failure, sustained ventricular tachycardia (VT), ventricular fibrillation (VF), and death were documented in the HCM Registry. All-cause mortality statistics for our study population were obtained by linking our database to the Social Security Death Index. A detailed description of methods is provided in the Supplemental Appendix S1, Section A2.

**Statistics.** Descriptive statistics were performed on patient demographics, hemodynamics, echocardiographic and CMR parameters, and cardiovascular events, stratified by the presence/absence of AF. Continuous variables are presented as mean ± standard deviation and categorical variables as the total number and percentage. Comparisons between patients with/without AF was performed using the independent $t$ test for continuous variables and the Fischer exact test for categorical variables. Statistical analyses were performed using STATA 14 (StataCorp LP, College Station, Texas).

## Computational methods

Please see Supplemental Appendix S1, Section B for detailed computational methods.

HCM patients with at least one episode of AF, either prior to their first clinic visit or during follow-up, were considered AF cases, and the remaining patients who were in sinus rhythm were labeled as No-AF.

Figure 1 summarizes the computational framework (HCM-AF-Risk Model) that we introduce for identifying HCM patients with AF. It comprises 5 steps: (1) pre-processing to remove select variables and address missing data; (2) feature selection, in which informative, predictive clinical variables that distinguish AF cases from No-AF are identified; (3) association analysis to quantify the degree of association between each predictor variable and the AF class; (4) supervised machine learning for building and training classifiers, and performing classification; and (5) quantitative and qualitative evaluation of the classifier's performance.

**Preprocessing.** We first removed variables that had no relevance to risk of AF (eg, visit date, patient ID), as well as variables representing adverse outcomes (ventricular tachycardia/fibrillation, heart failure, AF, stroke). The feature set remaining at the end of this step consisted of 93 clinical variables (Supplemental Table S1). As some of the records did not include values for all variables, data imputation was performed using a nearest-neighbor approach.

**Feature selection.** Feature selection was performed by assessing individual variables one at a time. We used the In-formation Gain criterion[18] to select highly predictive nominal attributes, and the 2-sample $t$ test under unequal variance[19,20] to select continuous features. Our feature selection process resulted in 18 clinical variables deemed to be informative and predictive of AF in HCM patients (Table 1).

**Association analysis.** To assess and express the degree and direction of association between the 18 predictor variables and the outcome variable, AF, we employed the polychoric cor-relation[21,22] which is applicable to both nominal and continuous variables. The polychoric correlation takes on values in the range [−1,1], where a negative value indicates negative association and a positive value corresponds to positive association (Table 1).

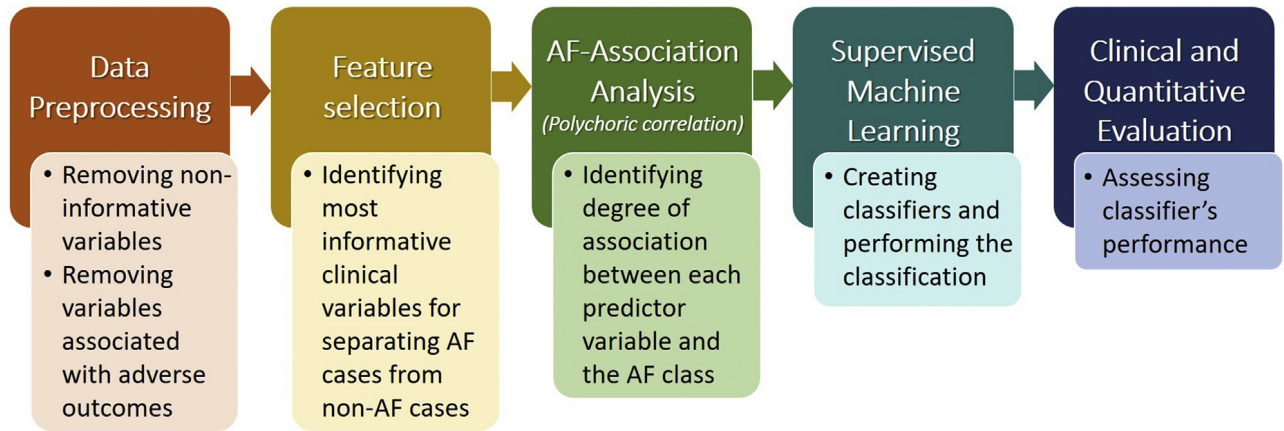

**Figure 1.** Schematic illustrating the Hypertrophic Cardiomyopathy Atrial Fibrillation (HCM-AF)-Risk Model.

**Supervised machine learning (classification).** Our classifier operates by taking as input a vector of values representing a patient's record and assigning a probability that indicates the patient's likelihood to belong to the AF vs No-AF class. The classifier calculates, for each 18-dimensional vector representing each patient, its probability of being an AF case vs its probability of being a No-AF case. The higher the value, the more likely the patient is to have AF.

To address the data imbalance resulting from the higher number of No-AF cases compared to AF cases in our cohort (No-AF:AF ratio of ∼3:1), we applied a combination of under- and oversampling (Fig. 2). Our method combines oversampling and undersampling, along with an ensemble of logistic regression and naïve Bayes classifiers, to address data imbalance[23] and separate AF records from No-AF records.

We employed a 5-fold cross-validation scheme to train and test our ensemble classifier. We partitioned the dataset into 5 equal subsets, 4 of which (80%) were used in turn as the training set; the 5th (20%) was left out and used for testing. The training/testing procedure was repeated 5 times—each time, a different subset was left out for testing. We applied the combined oversampling and undersampling approach to 80% of the dataset used for training, thus obtaining a balanced set over which to train the classifier. We obtained a balanced training set by applying our model only to the training dataset. We evaluated the performance of the trained model

**Table 1.** Eighteen variables identified as most informative for AF by our feature-selection method

| Variable | Variable type | $P$ (association with AF) | Polychoric correlation (association with AF) |
|---|---|---|---|
| Left atrial diameter, cm (+) | Continuous | 0.00000000001 | 0.316 |
| Heart rate at peak stress, bpm (−) | Continuous | 0.0000000001 | −0.288 |
| Age, y (+) | Continuous | 0.0000004 | 0.219 |
| Exercise metabolic equivalents (−) | Continuous | 0.0000014 | −0.154 |
| Septal myectomy (+) | Nominal | 0.0000021 | 0.353 |
| Exercise time, s (−) | Continuous | 0.0000031 | −0.225 |
| Diuretic treatment (+) | Nominal | 0.0000046 | 0.251 |
| Percentage of max heart rate achieved at peak exercise, % (−) | Continuous | 0.00058 | −0.156 |
| Heart rate recovery at 1 min post-exercise, bpm (−) | Continuous | 0.00072 | −0.205 |
| LV-LGE on CMR (presence +) | Nominal | 0.00092 | 0.269 |
| E/A (+) | Continuous | 0.001 | 0.091 |
| NYHA functional class (+) | Nominal | 0.0032 | 0.205 |
| E/e′ (+) | Continuous | 0.0033 | 0.157 |
| LV global longitudinal peak systolic strain rate, 1/s (+) | Continuous | 0.029 | 0.120 |
| Dyspnea on exertion (presence +) | Nominal | 0.035 | 0.198 |
| ABPR during exercise test in follow-up visit (+) | Continuous | 0.051 | 0.156 |
| Diastolic blood pressure at peak exercise, mm Hg (−) | Continuous | 0.053 | −0.105 |
| LV global longitudinal early diastolic strain rate, 1/s (−) | Continuous | 0.056 | −0.106 |

ABPR, abnormal blood pressure response; bpm, beats per minute; AF, atrial fibrillation; CMR, cardiac magnetic resonance imaging; E/A, ratio of early diastolic mitral flow velocity to the late diastolic mitral flow velocity; E/e′, ratio of early diastolic mitral flow velocity to the early diastolic mitral septal annulus motion velocity; LGE, late gadolinium enhancement; LV, left ventricle; NYHA: New York Heart Association. .

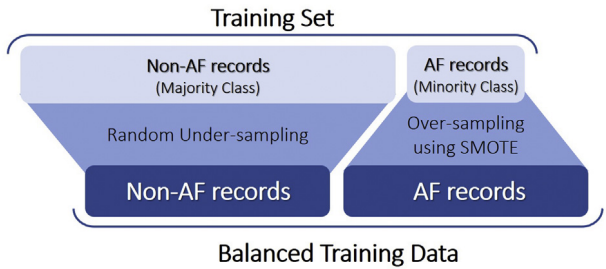

Figure 2. Methods for addressing data imbalance—the illustration shows our classification scheme for combining oversampling and undersampling. The topmost layer represents the entire training set, which comprises a majority of No-atrial fibrillation (AF) records (left) and the minority of AF records (right). The majority class in the training set (No-AF) is randomly undersampled such that the No-AF to AF record ratio is 2:1. The minority class (AF) is oversampled using Synthetic Minority Oversampling Technique (SMOTE) to generate synthetic new AF-like records, doubling the number of AF records. The resulting set forms a balanced training set, containing the same number of AF and No-AF records.

on the *imbalanced* left-out test set comprising 20% of the data. The classifier assigned to each record in the test set a probability to be associated with AF or non-AF. We conducted 10 complete 5-fold cross-validation experiments (each using a different 5-way split of the dataset), for a total of 50 complete runs, on 50 training and test sets.

**Model evaluation.** We employed specificity, sensitivity, and area under the receiver operating characteristics (ROC) curve to assess model performance.

**Comparison.** We evaluated the performance attained by our model when it was trained on 3 additional feature sets reported as being predictive of AF in the general population, namely the Framingham Heart Study (FHS),[24] the Atherosclerosis Risk in Communities (ARIC) study,[25] and the Cohorts of Heart and Aging Research in Genomic Epidemiology (CHARGE-AF) Consortium.[26]

## Results

### Patient population

We studied 831 patients with a clinical diagnosis of HCM. AF was diagnosed in 22% of the HCM cohort: 139 patients were diagnosed with AF prior to/at the first clinic visit, and 52 patients were diagnosed with AF during follow-up (Fig. 3). The prevalence of AF varied from 9% to 30% and increased with age; AF prevalence was highest in the age group of 61-80 years (Fig. 4).

Demographic, clinical, and imaging features of the HCM cohort at the time of their first clinic visit are presented in Table 2. Patients in the AF group were older, and were more likely to have higher New York Heart Association (NYHA) class, and lower exercise capacity, than the No-AF group. The AF group also had larger left atrial (LA) size (Fig. 5), greater diastolic dysfunction, worse global longitudinal strain, and a greater amount of LV replacement fibrosis (reflected by LV-LGE), compared with the No-AF group, suggesting a greater degree of LV myopathy. No difference was observed in LV mass, maximum LV thickness, or left ventricular outflow tract (LVOT) gradients between the AF and No-AF groups of HCM patients.

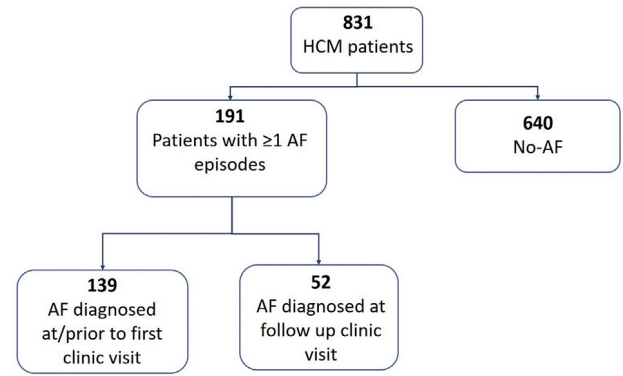

Figure 3. Flow chart indicating selection of atrial fibrillation (AF) and No-AF cases in the hypertrophic cardiomyopathy (HCM) cohort.

Mean follow-up was 3.1 years (median = 2.1; 25th–75th percentile = 1.0–4.8 years). HCM patients from the AF group had a higher incidence of heart failure and all-cause death, compared with the No-AF group (Table 2).

### Machine learning–based identification of AF cases

Our feature-selection process identified 18 clinical variables whose values distinguish AF cases from No-AF cases within the HCM population. Table 1 provides a list of these predictive variables, along with the corresponding polychoric correlation and $P$ values, indicating their degree of association (or lack thereof) with AF. We identified 7 variables that are negatively correlated with AF, and 11 variables that are positively associated with AF. Left atrial diameter is highly correlated with AF. Several exercise-related parameters, including, lower exercise capacity (reflected by lower metabolic equivalents [METs], exercise time, peak stress heart rate), abnormal blood pressure (BP) response to exercise, lower diastolic BP at peak exercise, and lower heart rate recovery after exercise are associated with higher risk for AF. Other predictors of AF include replacement fibrosis in the LV (reflected by LV-LGE), greater diastolic dysfunction (reflected

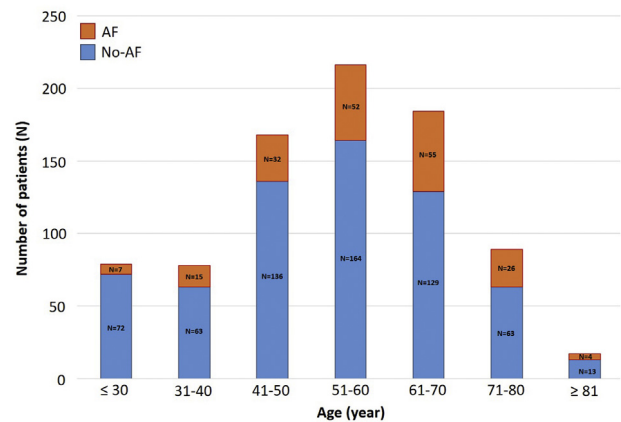

Figure 4. Age distribution of hypertrophic cardiomyopathy (HCM) patients with atrial fibrillation (AF) in HCM cohort. AF prevalence increases with age in HCM.

**Table 2.** Demographic and clinical feature values of the HCM cohort shown for patients with/without AF

| Variable | No-AF n = 640 | AF n = 191 | P |
|---|---|---|---|
| **Clinical characteristics** | | | |
| Age, y | 52 ± 16 | 58 ± 13 | <0.001 |
| Male sex | 399 (62) | 120 (63) | 0.9 |
| Body mass index, kg/m² | 29 ± 6 | 30 ± 7 | 0.1 |
| HCM type | | | 0.02 |
|   Non-obstructive | 203 (32) | 56 (30) | |
|   Labile obstructive | 244 (38) | 57 (30) | |
|   Obstructive | 191 (30) | 77 (41) | |
| NYHA class | | | <0.001 |
|   I | 376 (59) | 82 (43) | |
|   II-III | 264 (41) | 109 (57) | |
| Angina | 260 (41) | 68 (36) | 0.2 |
| Family history of HCM | 125 (20) | 37 (19) | 0.9 |
| ICD implantation | 42 (7) | 32 (17) | <0.001 |
| Syncope | 120 (19) | 41 (22) | 0.5 |
| Family history of sudden cardiac death | 159 (25) | 44 (23) | 0.7 |
| Non-sustained VT | 62 (10) | 29 (15) | 0.05 |
| Septal wall thickness ≥30 mm | 55 (9) | 8 (4) | 0.07 |
| Medications | | | |
|   Beta-blocker | 435 (68) | 157 (82) | <0.001 |
|   Calcium channel blocker | 176 (28) | 66 (35) | 0.07 |
|   RAS blockade | 150 (23) | 49 (26) | 0.6 |
|   Disopyramide | 15 (2) | 15 (8) | 0.001 |
| **Imaging features** | | | |
| Echocardiography | | | |
|   Left atrial diameter, mm | 41 ± 7 | 45 ± 8 | <0.001 |
|   Maximal septal wall thickness, mm | 21 ± 5 | 21 ± 5 | 0.1 |
|   LV ejection fraction, % | 66 ± 8 | 64 ± 8 | 0.1 |
|   E/A | 1.3 ± 0.6 | 1.7 ± 1.3 | <0.001 |
|   E/é | 18 ± 11 | 22 ± 12 | <0.001 |
|   Rest LVOT peak gradient, mm Hg | 28 ± 32 | 32 ± 30 | 0.2 |
|   Stress LVOT peak gradient, mm Hg | 67 ± 53 | 72 ± 54 | 0.3 |
|   LV-GLS, % | −16.1 ± 3.7 | −15.1 ± 3.8 | 0.003 |
|   LV-SR_S | −1.00 ± 0.20 | −0.93 ± 0.20 | <0.001 |
|   LV-SR_E | 1.14 ± 0.35 | 1.06 ± 0.31 | <0.001 |
|   Moderate or severe MR | 70 (11) | 31 (16) | 0.06 |
| Cardiac magnetic resonance (n = 608) | | | |
|   LV mass, g | 163 ± 69 | 172 ± 62 | 0.2 |
|   LGE presence | 300 (63) | 107 (81) | <0.001 |
|   LGE (% of LV mass) | 11 ± 12 | 16 ± 13 | 0.004 |
| Positron emission tomography (n = 145) | | | |
|   Global rest MBF, mL/min/g | 0.93 ± 0.26 | 1.04 ± 0.54 | 0.2 |
|   Global stress MBF, mL/min/g | 2.14 ± 0.65 | 2.01 ± 0.75 | 0.3 |
|   Myocardial flow reserve | 2.44 ± 0.85 | 2.30 ± 0.66 | 0.3 |
|   Summed difference scores | 5.4 ± 4.9 | 4.3 ± 4.8 | 0.2 |
| Exercise parameters | | | |
|   Treadmill exercise time, s | 566 ± 202 | 471 ± 193 | <0.001 |
|   Metabolic equivalents | 10.3 ± 4.2 | 8.8 ± 3.7 | <0.001 |
|   Rest heart rate, bpm | 65 ± 13 | 65 ± 14 | 0.5 |
|   Rest systolic blood pressure, mm Hg | 133 ± 22 | 131 ± 18 | 0.5 |
|   Rest diastolic blood pressure, mm Hg | 77 ± 11 | 77 ± 12 | 0.9 |
|   Stress heart rate, bpm | 145 ± 28 | 131 ± 26 | <0.001 |
|   Stress systolic blood pressure, mm Hg | 161 ± 36 | 153 ± 37 | 0.02 |
|   Stress diastolic blood pressure, mm Hg | 82 ± 18 | 77 ± 17 | 0.003 |
|   ABPR | 179 (31) | 70 (43) | 0.009 |
| **Follow-up: adverse outcomes** | | | |
| Stroke | 7 (1) | 5 (3) | 0.2 |
| Heart failure hospitalization | 27 (5) | 16 (10) | 0.04 |
| VT/VF | 18 (3) | 11 (7) | 0.1 |
| Death | 13 (2) | 13 (8) | 0.003 |

Values are n (%), unless otherwise indicated.

ABPR, abnormal blood pressure response to exercise; AF, atrial fibrillation; bpm, beats per minute; E/A, ratio of early diastolic mitral flow velocity to the late diastolic mitral flow velocity; E/e', ratio of early diastolic mitral flow velocity to the early diastolic mitral septal annulus motion velocity; HCM, hypertrophic cardiomyopathy; ICD, implantable cardioverter defibrillator; LGE, late gadolinium enhancement; LV, left ventricular;; LV-GLS, LV peak global longitudinal systolic strain; LVOT, LV outflow tract; LV-SR_E, LV peak global longitudinal early diastolic strain rate; LV-SR_S, peak global longitudinal systolic strain rate; MBF, myocardial blood flow; MR, mitral regurgitation; NYHA: New York Heart Association; RAS blockade, angiotensin-converting enzyme inhibitor, angiotensin II receptor blocker; VT/VF, ventricular tachycardia/ventricular fibrillation.

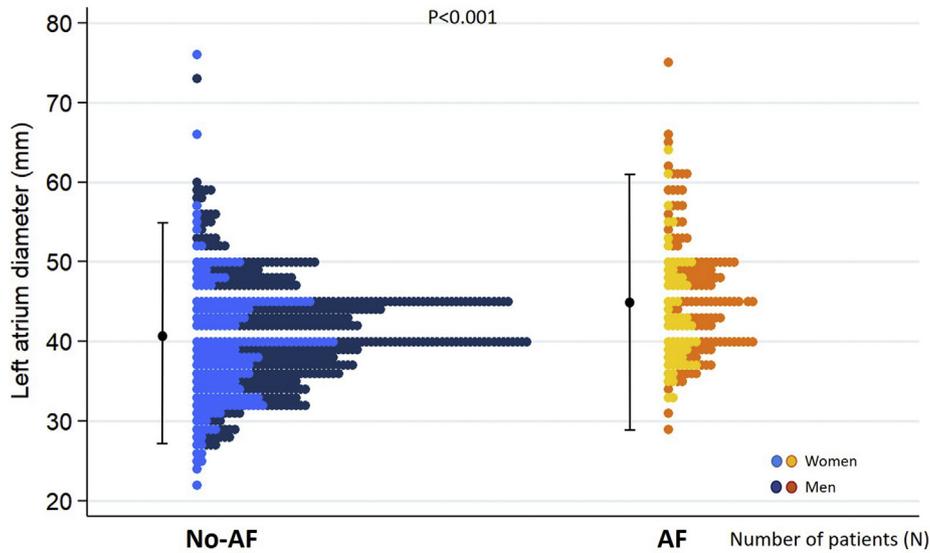

**Figure 5.** Distribution of left atrium (LA) size in hypertrophic cardiomyopathy patients with/without atrial fibrillation (AF). Significant overlap exists in LA diameter values in the AF and No-AF groups, but mean values for LA diameter were significantly higher in the AF group, compared with the No-AF group ($P < 0.001$). Each dot represents a patient; mean ± 1.96 standard deviations is presented.

by higher ratio of early diastolic mitral flow velocity to the late diastolic mitral flow velocity [E/A] and ratio of early diastolic mitral flow velocity to the early diastolic mitral septal annulus motion velocity [E/e'], and lower left ventricular peak global longitudinal early diastolic strain rate [LV-SR_E]) and worse (more positive) global longitudinal systolic strain rate (LV-SR_S).

Notably, combining the ensemble classifier comprising logistic regression and naïve Bayes with oversampling and undersampling led to higher sensitivity and area under the receiver operating curve (AUC), compared to the 4 simple classifiers (naïve Bayes, logistic regression, decision tree, and random forest) alone (Table 3). Figure 6 illustrates the C-index (0.80) for our method (HCM-AF-Risk Model), which assigns an individualized probability to each patient who presents with AF.

## Comparison of HCM-AF-Risk Model performance with previous AF models

We compared the performance of the HCM-AF-Risk Model in identifying AF cases with that obtained using

**Table 3.** Comparison of performance between the simple baseline logistic regression classifier (Baseline), approaches used for addressing data imbalance (Random undersampling), and the classifier resulting from our combination of undersampling and oversampling, trained on datasets represented via the 18 features identified by our feature selection method (HCM-AF-Risk Model)

| Performance measure | Baseline | Random undersampling | HCM-AF-Risk Model |
|---|---|---|---|
| Sensitivity | 0.41 (±0.04) | 0.43 (±0.04) | **0.74** (±0.02) |
| Specificity | **0.93** (±0.03) | 0.90 (±0.02) | 0.70 (±0.03) |
| AUC (C-index) | 0.79 (±0.04) | 0.77 (±0.02) | **0.80** (±0.03) |

Standard deviation is shown in parentheses; highest values are shown in boldface.

AUC, area under the receiver operating characteristic curve; HCM-AF, hypertrophic cardiomyopathy atrial fibrillation.

features employed by the FHS,[24] ARIC,[25] and CHARGE-AF[26] risk models (Table 4). The HCM-AF-Risk Model demonstrates significantly higher performance ($P < 0.001$) across all evaluation metrics, including specificity, sensitivity, and area under ROC curve (C-index) for identification of HCM patients with AF, compared with published models[24-26] focused on AF prediction in the general population. The datasets used in these studies are not publicly available, which precludes their use for training/testing on our dataset and comparing their performance according to all the measures we have used. Hence, we compared the performance level attained by our model with that reported by the other studies in terms of the C-index (Table 4).

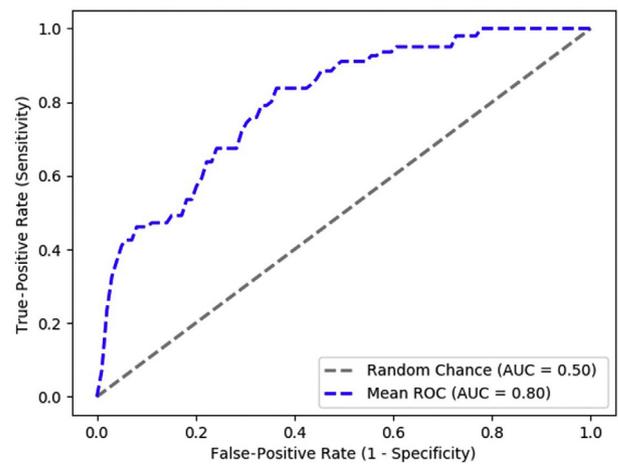

**Figure 6.** Receiver operating characteristic (ROC) curve for Hypertrophic Cardiomyopathy Atrial Fibrillation (HCM-AF)-Risk Model. The ROC curve depicts the performance of the HCM-AF-Risk Model that combines the undersampling and oversampling approaches. The false-positive rate is shown on the x-axis, and the true-positive rate is indicated on the y-axis. AUC, area under the curve.

**Table 4.** Comparison of performance attained by HCM-AF-Risk Model based on 4 feature sets

| Feature set #/originating study, features | Sensitivity | Specificity | C-index/AUC |
|---|---|---|---|
| 1/FHS<br>Age, sex, body mass index, systolic blood pressure at rest, treatment for hypertension, heart failure | 0.53 (±0.20) | 0.60 (±0.20) | 0.60 (±0.20) |
| 2/ ARIC<br>Age, race, height, smoking status, systolic blood pressure, hypertension medication use, left atrial enlargement by echocardiography, diabetes, coronary artery disease, heart failure | 0.57 (±0.05) | 0.63 (±0.02) | 0.68 (±0.05) |
| 3/CHARGE-AF Consortium<br>Age, race, height, weight, systolic blood pressure, diastolic blood pressure, current smoking, antihypertensive medication use, diabetes, history of myocardial infarction, history of heart failure | 0.54 (±0.10) | 0.60 (±0.10) | 0.61 (±0.10) |
| 4/HCM-AF-Risk Model (our study)<br>18 features (shown in Table 1) identified by our feature-selection method | **0.74** (±0.02) | **0.70** (±0.03) | **0.80** (±0.03) |

The variables associated with the 4 feature sets that were used to represent the data for training our model are indicated in italics in this footnote. Set 1 shows our model performance when trained on 6 attributes identified as informative for AF prediction in the study by Schnabel et al.[24] (C-index/AUC = 0.78), conducted using the FHS dataset; the predictors *PR interval by EKG* and *significant cardiac murmur* were not recorded in our dataset. Set 2 shows the performance attained by our model when trained on 10 features reported as predictive of AF in the study by Chamberlain et al.[25] using the ARIC dataset (C-index/AUC = 0.76)[5]; the predictors *precordial murmur* and *LVH by EKG* were not recorded in our dataset. Set 3 includes 11 risk factors for AF identified in the study by Alonso et al.[26] in the CHARGE-AF Consortium (C-index/AUC = 0.76). Set 4 shows the performance achieved by our model based on the 18 features identified as predictive by our feature-selection approach. The difference between the performance attained when the representation is based on our feature set (Set 4) and those attained when the representation is based on the 3 other sets, is highly statistically significant ($P < 0.001$). Standard deviation is shown in parentheses. The highest values are shown in boldface.

AF, atrial fibrillation; ARIC, Atherosclerosis Risk in Communities Study; AUC, area under the curve; CHARGE-AF, Cohorts for Heart and Aging Research in Genomic Epidemiology-Atrial Fibrillation; EKG, electrocardiogram; FHS, Framingham Heart Study; HCM-AF, hypertrophic cardiomyopathy atrial fibrillation; PR interval, the time from the onset of the P wave to the start of the QRS complex on electrocardiogrpahy; LVH, left ventricular hypertrophy.

We repeated our experiments, using only LA diameter to represent our dataset, based on results from several previous studies[2,27,28] that identified LA enlargement as most predictive of AF in HCM. We observed reduction in our model's performance when LA diameter alone is included in the feature set: specifically, area under ROC curve decreased by 20% (C-index: 0.66 down from 0.80); sensitivity decreased by 37% (0.54 down from 0.74); and specificity decreased by 14% (0.63 down from 0.70).

## Discussion

The HCM-AF-Risk Model is the first machine learning—based method for the identification of AF cases and clinical/imaging features associated with higher/lower risk of AF in HCM, using electronic health record data. In our model, individual patient data are represented as an N-dimensional vector, and the model output is a probability score for AF (AF risk) in HCM. We identified 18 clinical variables that are highly associated (positively/negatively) with AF in HCM patients. In addition to age, NYHA class, LA size, and LV fibrosis, which have been previously found to be associated with AF in HCM,[29-31] we found that additional clinical features, such as LV diastolic dysfunction and lower LV-systolic strain, are positively associated with AF, and greater exercise capacity is negatively associated with AF in HCM.

Our machine learning model directly addresses the imbalance inherent in clinical data—a relatively small proportion of the patients present with an adverse outcome, AF in our case, and the majority do not exhibit this condition (the rest of the HCM population in this study). Handling such imbalance is critical when using machine learning to obtain a classifier. Otherwise, the classifier learned based on the data tends to favor assigning patients to the negative class, which is in the majority (No-AF).

Several small studies have reported an association between age,[29] NYHA class,[30] left atrial (LA) size/function,[29-31] EKG-P-wave dispersion,[31] N-terminal proB-type natriuretic peptide (NT-proBNP) levels,[31] fibrosis in the LV,[30] and AF in HCM patients. Clinical risk scores for AF prediction have been developed and validated using general populations from the FHS,[24] the Cardiovascular Heart Study (CHS),[26] ARIC,[25] the Multi-Ethnic Study of Atherosclerosis Study (MESA),[32] the Reykjavik Study (AGES),[33] and the Rotterdam Study (RS).[34] But it is unknown whether these models are effective in assessing AF risk in HCM, given the differences in cardiac physiology/pathology between HCM patients and the general population. In our retrospective study, we devised a machine learning—based model to *identify* HCM patients with AF/history of AF; the next step is prospective multicentre testing and validation of the HCM-AF-Risk Model to predict AF in HCM.

**Table 5.** Comparison of clinical/imaging features associated with AF, VARs, and HF in HCM patients, identified by the HCM-AF-Risk Model (current work), the HCM-VAr-Risk Model,[23] and the HCM-HF-Risk Model

| AF<br>HCM-AF-Risk Model (18 predictive variables)<br>sensitivity = 0.74, specificity = 0.70,<br>C-index = 0.80 | | | VAr HCM-VAr-Risk Model (22 predictive variables)<br>sensitivity = 0.73, specificity = 0.76,<br>C-index = 0.83 | | | HF<br>HCM-HF-Risk Model (17 predictive variables)<br>sensitivity = 0.80, specificity = 0.78,<br>C-index = 0.84 | | |
|---|---|---|---|---|---|---|---|---|
| Variables associated with AF in HCM | $P$ | Polychoric correlation with AF | Variables associated with VAr (VT/VF) in HCM | $P$ | Polychoric correlation with VAr | Variables associated with HF in HCM | $P$ | Polychoric correlation with HF |
| Exercise time, s (−) | $3 \times 10^{-6}$ | −0.225 | Exercise time, s (−) | $7 \times 10^{-2}$ | −0.167 | Exercise time, s (−) | $4.7 \times 10^{-7}$ | −0.346 |
| Exercise METs (−) | $1 \times 10^{-6}$ | −0.154 | Exercise METs (−) | $1 \times 10^{-2}$ | −0.131 | Exercise METs (−) | $< 1 \times 10^{-9}$ | −0.579 |
| Age, y (+) | $4 \times 10^{-7}$ | 0.219 | Age, y (−) | $3 \times 10^{-2}$ | −0.15 | Sex (male +) | $> 1 \times 10^{-9}$ | 0.401 |
| E/e′ (+) | $3 \times 10^{-3}$ | 0.157 | E/e′ (+) | $6 \times 10^{-2}$ | 0.167 | LV-LGE % of LV mass (+) | $3 \times 10^{-2}$ | 0.191 |
| LV global longitudinal peak systolic strain rate, 1/s (+) | $2.9 \times 10^{-2}$ | 0.12 | LV global longitudinal peak systolic strain rate, 1/s (+) | $3 \times 10^{-3}$ | 0.171 | History of syncope (+) | $4.7 \times 10^{-2}$ | 0.157 |
| LV global longitudinal early diastolic strain rate, 1/s (−) | $5 \times 10^{-2}$ | −0.106 | LV global longitudinal early diastolic strain rate, 1/s (−) | $1 \times 10^{-3}$ | −0.213 | History of smoking (+) | $2.5 \times 10^{-2}$ | 0.148 |
| HR at peak exercise stress, bpm (−) | $1 \times 10^{-10}$ | −0.288 | NSVT (presence +) | $5 \times 10^{-4}$ | 0.994 | HR at peak exercise stress, bpm (−) | $< 1 \times 10^{-9}$ | −0.447 |
| LV-LGE (presence +) | $9 \times 10^{-4}$ | 0.269 | VT induced by NIPS during follow-up (presence +) | $1 \times 10^{-2}$ | 0.667 | LV-LGE (presence +) | $3.6 \times 10^{-2}$ | 0.159 |
| HRR at 1 min post-exercise, bpm (−) | $7 \times 10^{-4}$ | −0.205 | HCM type (non-obstructive +) | $1 \times 10^{-3}$ | 0.366 | HRR at 1 min post-exercise, bpm (−) | $< 1 \times 10^{-9}$ | −0.411 |
| Dyspnea on exertion (+) | $3 \times 10^{-2}$ | 0.198 | Peak stress LVOT gradient, mm Hg (−) | $1 \times 10^{-5}$ | −0.273 | Dyspnea on exertion (+) | $< 1 \times 10^{-9}$ | 0.668 |
| % of max HR at peak exercise, % (−) | $5 \times 10^{-4}$ | −0.156 | Unexplained syncope (presence +) | $3 \times 10^{-4}$ | 0.264 | % of max HR at peak exercise, % (+) | $1 \times 10^{-7}$ | 0.546 |
| Septal myectomy (+) | $2.1 \times 10^{-6}$ | 0.353 | LV global longitudinal peak systolic strain, % (+) | $3 \times 10^{-2}$ | 0.235 | Presyncope (+) | $4.2 \times 10^{-2}$ | 0.146 |
| Left atrial diameter, cm (+) | $1 \times 10^{-11}$ | 0.316 | SBP before exercise test, mm Hg (−) | $1 \times 10^{-3}$ | −0.232 | Late diastolic filling velocity (A), cm/s (+) | $6 \times 10^{-3}$ | 0.136 |
| Diuretic treatment (+) | $4.6 \times 10^{-6}$ | 0.251 | ECHO LVEF, % (−) | $1 \times 10^{-2}$ | −0.198 | Family history of HCM (−) | $4.9 \times 10^{-2}$ | −0.132 |
| NYHA functional class (+) | $3 \times 10^{-3}$ | 0.205 | Family history of HCM (presence +) | $6 \times 10^{-2}$ | 0.195 | LV end-diastolic volume, ml (−) | $2.3 \times 10^{-2}$ | −0.111 |
| ABPR during exercise test in follow-up visit (presence +) | $5 \times 10^{-2}$ | 0.156 | IVS/PW ratio (+) | $1 \times 10^{-2}$ | 0.195 | LV end systolic volume, ml (−) | $2.7 \times 10^{-2}$ | −0.096 |
| DBP at peak exercise, mm Hg (−) | $5 \times 10^{-2}$ | −0.105 | DBP before exercise test, mm Hg (−) | $1 \times 10^{-2}$ | −0.177 | Peak stress LVOT gradient, mm Hg (+) | $4.8 \times 10^{-2}$ | 0.0714 |
| E/A (+) | $1 \times 10^{-3}$ | | Maximal IVS thickness, mm (+) | $3 \times 10^{-3}$ | 0.125 | | | |
| | | | Peak rest LVOT gradient, mm Hg (−) | $4 \times 10^{-2}$ | −0.119 | | | |
| | | | Body mass index, kg/m² (−) | $3 \times 10^{-2}$ | −0.115 | | | |
| | | | Family history of SCD (presence +) | $5 \times 10^{-2}$ | 0.097 | | | |
| | | | Statin use (−) | $6 \times 10^{-2}$ | −0.052 | | | |

The HCM-AF-Risk Model (current work), the HCM-VAr-Risk Model,[23] and the HCM-HF-Risk Model were developed using similar methods and the same HCM patient dataset.

ABPR, abnormal blood pressure response; AF, atrial flutter or atrial fibrillation of any duration before 1st clinic visit and/or during follow up; bpm, beats per minute; DBP, diastolic blood pressure; E/A: ratio of early diastolic mitral flow velocity to the late diastolic mitral flow velocity; ECHO, echocardiogram; E/e′: ratio of early diastolic mitral flow velocity to the early diastolic mitral septal annulus motion; HCM, hypertrophic cardiomyopathy; HF, heart failure (≥NYHA class III symptoms and/or HF hospitalization during follow-up); HR, heart rate; HRR, heart rate recovery; IVS, interventricular septum; IVS/PW, ratio of maximal thickness of interventricular septum and maximal thickness of posterior wall of left ventricle; LV, left ventricle; LVEF, LV ejection fraction; LV-LGE, late gadolinium enhancement in the LV myocardium by cardiac magnetic resonance imaging; LVOT, LV outflow tract; MET, metabolic equivalent; NIPS, non-invasive programmed stimulation; NSVT, non-sustained ventricular tachycardia; NYHA, New York Heart Association; SBP, systolic blood pressure; SCD, sudden cardiac death; VAr, sustained ventricular tachycardia (≥ 30 s) or ventricular fibrillation before 1st clinic visit and/or during follow-up; VF, ventricular fibrillation; VT, ventricular tachycardia.

## HCM-AF-Risk Model

Employing a statistical machine learning method is advantageous as it allows automatic quantification of the likelihood of an event (AF, in this case) based on the combination of feature values obtained from the patient's electronic health records, and their level of association with the event. Moreover, unlike traditional rule-based models,[5] machine learning methods are flexible in the face of new data, as these methods can tune and update the parameters that govern the classification algorithm based on the added data. Thus, machine learning methods are well suited for use in the clinical setting, in which additional patients' data are frequently accumulated.

In contrast to the majority of current machine learning methods based on artificial neural networks,[6,35,36] for which the output decision typically cannot be explained, our method is based on modeling a clear probabilisitic decision process that can be tracked back and used to justify the decision. We believe that this aspect of machine learning is critical when it is used to support clinical decision making. Notably, our HCM-AF-Risk Model addresses data imbalance, and it utilizes a set of 18 clinical variables to identify AF cases, and clinical features associated with higher/lower risk for AF in HCM patients.

We note that heart failure, along with VT/VF and stroke, was not included in the list of clinical variables considered by our method. is the reason it was excluded is that our goal is to identify demographic, clinical, and imaging features that predict adverse outcomes (AF in this case) in HCM patients, and using such adverse outcomes as predictors defeats this purpose.

### Clinical predictors of AF in HCM using the HCM-AF-Risk Model

Left atrial diameter is the strongest predictor of AF in our study. The association between LA size and AF has been documented extensively in the general population[37-41] and in HCM patients.[2,42-46] The association between LA enlargement and AF has been attributed to stretch-induced LA structural and electrophysiologic remodeling.[47] In the case of HCM, given that most causal HCM mutations are expressed in both atrial and ventricular myocytes, atrial myopathy and LV diastolic dysfunction could underlie the high prevalence of AF in HCM.

Our HCM-AF-Risk Model indicates an association between diastolic dysfunction and AF in HCM. We found that higher values for E/A, E/e′,[48] and lower (worse) global diastolic strain rate reflecting a greater degree of diastolic dysfunction are associated with higher risk for AF in HCM. Similar results have been reported in studies conducted in non-HCM patients.[37,49,50] The mechanism whereby diastolic dysfunction has been proposed to predispose patients to AF is by increasing LA preload (stretch), afterload, and wall stress (dilation), which lead to ion channel remodeling and fibrosis and increase susceptibility for reentrant arrhythmias such as atrial fibrillation/flutter.[49]

LV-LGE and worse LV global longitudinal peak systolic strain rate, which reflect a greater degree of LV myopathy, are associated with AF in our model. Several studies previously have detected an association between LV fibrosis and AF in HCM.[51-53] A recent CMR study in HCM patients reported greater amounts of LA fibrosis in HCM patients with PAF, as well as a positive association between atrial and ventricular fibrosis (LGE).[54] Since fibrosis slows conduction and predisposes to reentry, LA fibrosis would be expected to increase risk for AF.

Lower exercise capacity, lower chronotropic response/heart rate recovery, abnormal BP response to exercise, and lower diastolic BP at peak exercise are associated with higher risk for AF in our study. Similar results of exercise intolerance in HCM patients with PAF have been reported in a previous study of 265 HCM patients during sinus rhythm[55]—in this case, the authors did not observe an association between lower exercise capacity and diastolic dysfunction or LA volume. Additionally, ECHO[56] and CMR[29,54] studies in HCM patients have revealed greater impairment of LA function and a greater degree of LA fibrosis in HCM patients with PAF, suggesting that PAF is a marker of LA myopathy.

One mechanism underlying reduced exercise capacity in HCM patients (with PAF), even during sinus rhythm,[55] could be impairment of LV hemodynamics in the setting of LA myopathy, since the LA modulates LV performance by its reservoir function during ventricular systole, conduit function during early ventricular diastole, and booster pump function during late ventricular diastole. A second possibility is higher pulmonary capillary wedge pressure in HCM patients with AF, based on results of a study in 123 patients who underwent simultaneous left and right heart catheterization; pulmonary capillary wedge pressure was higher than LV end-diastolic pressure among AF patients and lower than LV end-diastolic pressure among patients in sinus rhythm.[56] Other contributors to lower exercise capacity in HCM patients with AF include sympathovagal imbalance[57] leading to systemic vasodilation, chronotropic incompetence induced by atrial remodeling/medications, and greater degree of LV myopathy.

### Comparison of clinical variables associated with atrial fibrillation, VT/VF, and heart failure in HCM.

We have also developed machine learning—based models for identifying HCM patients with lethal ventricular arrhythmias (HCM-VAr-Risk Model)[23] and heart failure (HCM-HF-Risk Model; Table 5) using the same methodology and dataset used to generate the HCM-AF-Risk Model. Comparison of these model results revealed 2 predictors that are common to all 3 models (exercise time, exercise METs).

We identified 5 predictors (exercise time, METs, E/e′ ratio, LV global longitudinal peak systolic strain rate, and LV global longitudinal early diastolic strain rate) that are common in the AF and VAr models; 13 variables are associated with AF, but not with VT/VF (Table 5). Older age is associated with increased risk for AF, but lower risk for VT/VF, which has been confirmed by other studies.[2,58] HCM type (nonobstructive), family history of HCM or sudden cardiac death, and non-sustained VT are associated with VT/VF but not AF, which may reflect differences in arrhythmic substrate in the LV and LA in HCM. Notably, LV hypertrophy (max interventricular septum thickness, interventricular septum /posterior wall ratio) is associated with VT/VF but not AF—higher risk for VT/VF but not AF could be attributed to a greater degree of myocardial ischemia,[59] interstitial fibrosis,[60] and myocyte disarray[61] in the hypertrophied LV.[62] The association of replacement fibrosis (LV-LGE) with AF but not VT/

VF could reflect the impact of a greater degree of diastolic dysfunction induced by LV fibrosis resulting in LA dilatation/remodeling and AF. Taken together, our results suggest distinct pathophysiologic mechanisms underlying atrial and ventricular arrhythmias in HCM (Table 5).

For heart failure (HF), 7 variables are negatively correlated with HF, and the remainder (n = 10) are positively associated with HF. We identified 7 common predictors between the AF and HF models (dyspnea on exertion, exercise-METs/time, heart rate at peak stress, percentage of maximum heart rate at peak exercise, heart rate recovery at 1 min post-exercise, and presence of LV-LGE on CMR). Lower values of LV end-diastolic volume, LV end-systolic volume, exercise capacity and higher values of stress LVOT gradients, and LV-LGE are positively associated with HF in HCM. These results suggest that LV geometry, obstruction, fibrosis, and diastolic dysfunction contribute to HF in HCM.

## Limitations

This is a single centre, retrospective study. Although our approach is especially effective when the size of the dataset and the number of examples in the underrepresented class (AF in our case) are limited, our approach has limitations. When working with a larger imbalanced dataset, the undersampling step involved in creating a balanced training set eliminates a sizeable portion of the overrepresented class, whereas the oversampling step applied via the **S**ynthetic **M**inority **O**versampling **T**echnique (SMOTE) process to the underrepresented class generates a large number of synthesized samples that were not in the original dataset. Both of these lead to a potential loss of useful information and alter the distribution of characteristic feature values across both the minority and majority classes. In the current study, this issue was mitigated by thorough experimentation to determine the effective rates of undersampling and oversampling.

We grouped all AF cases (paroxysmal, persistent, permanent), as well as prevalent and incident AF, into one set because of the low event number; hence, some of the risk markers of AF (eg, LA size) could reflect a consequence of AF. Furthermore, we only included symptoms and ECHO/CMR imaging features obtained at the patients' first clinic visit in the model. Given that symptoms and cardiac physiology can evolve over time, the relationship between evolution of individual features and AF deserves investigation. Hence, periodic reassessment of risk is needed in the clinical setting.

LA volume,[44] LA strain,[63] LA fibrosis,[54] EKG/blood biomarkers,[64] genotype,[65-67] and sleep apnea were not included in our model, because these data are not available for a large proportion of our cohort. Lastly, we were unable to assess the generalizability of our approach by applying our developed model to additional HCM patients—beyond the cross-validation study—due to the unavailability of data from other HCM cohorts reported in other studies. We plan to address the latter issue in a future prospective study.

We could not assess the effect of the parameters, stroke, VT/VF, or heart failure on model performance, because of the small number of patients with these outcomes at the time of patients' first clinic visit.

## Conclusions

The HCM-AF-Risk Model effectively identifies HCM patients with AF. Our model attains good performance (0.74: sensitivity; 0.70: specificity; C-index: 0.80) while addressing the imbalance between high-risk and low-risk cases that is inherent in most clinical data. The set of clinical attributes identified by our method as being indicative of AF, and serving to justify the severity level assigned by the classifier, includes several hitherto unidentified markers of AF in HCM, and suggests that HCM patients with AF have a more severe cardiac HCM phenotype.

## Funding Sources

This work was funded in part by the National Science Foundation (NSF) IIS EAGER grant #1650851, and the National Institutes of Health grants R01 LM012527, and U54 GM104941 (to H.S.), an award from the John Taylor Babbitt (JTB) foundation (Chatham, New Jersey), and startup funds from the University of California San Francisco, Division of Cardiology (to M.R.A.).

## Disclosures

The authors have no conflicts of interest to disclose.

## References

1. Camm CF, Camm AJ. Atrial fibrillation and anticoagulation in hypertrophic cardiomyopathy. Arrhythm Electrophysiol Rev 2017;6:63-8.

2. Olivotto I, Cecchi F, Casey SA, et al. Impact of atrial fibrillation on the clinical course of hypertrophic cardiomyopathy. Circulation 2001;104:2517-24.

3. Tsuda T, Hayashi K, Fujino N, et al. Effect of hypertrophic cardiomyopathy on the prediction of thromboembolism in patients with nonvalvular atrial fibrillation. Heart Rhythm 2019;16:829-37.

4. Maron BJ, Olivotto I, Bellone P, et al. Clinical profile of stroke in 900 patients with hypertrophic cardiomyopathy. J Am Coll Cardiol 2002;39:301-7.

5. Desai RJ, Wang SV, Vaduganathan M, Evers T, Schneeweiss S. Comparison of machine learning methods with traditional models for use of administrative claims with electronic medical records to predict heart failure outcomes. JAMA Netw Open 2020;3:e1918962.

6. Fernandez-Ruiz I. Artificial intelligence to improve the diagnosis of cardiovascular diseases. Nat Rev Cardiol 2019;16:133.

7. Zou L, Yu S, Meng T, et al. A technical review of convolutional neural network-based mammographic breast cancer diagnosis. Comput Math Methods Med 2019;2019:6509357.

8. Daoud M, Mayo M. A survey of neural network-based cancer prediction models from microarray data. Artif Intell Med 2019;97:204-14.

9. Lu DY, Ventoulis I, Liu H, et al. Sex-specific cardiac phenotype and clinical outcomes in patients with hypertrophic cardiomyopathy. Am Heart J 2019;219:58-69.

10. Gersh BJ, Maron BJ, Bonow RO, et al. 2011 ACCF/AHA guideline for the diagnosis and treatment of hypertrophic cardiomyopathy: a report of the American College of Cardiology Foundation/American Heart Association Task Force on Practice Guidelines. Circulation 2011;124:e783-831.

11. Sankaranarayanan R, Fleming EJ, Garratt CJ. Mimics of hypertrophic cardiomyopathy—diagnostic clues to aid early identification of phenocopies. Arrhythm Electrophysiol Rev 2013;2:36-40.

12. January CT, Wann LS, Alpert JS, et al. 2014 AHA/ACC/HRS guideline for the management of patients with atrial fibrillation: executive summary: a report of the American College of Cardiology/American Heart Association Task Force on practice guidelines and the Heart Rhythm Society. Circulation 2014;130:2071-104.

13. January CT, Wann LS, Alpert JS, et al. 2014 AHA/ACC/HRS guideline for the management of patients with atrial fibrillation: a report of the American College of Cardiology/American Heart Association Task Force on practice guidelines and the Heart Rhythm Society. Circulation 2014;130:e199-267.

14. Lang RM, Badano LP, Mor-Avi V, et al. Recommendations for cardiac chamber quantification by echocardiography in adults: an update from the American Society of Echocardiography and the European Association of Cardiovascular Imaging. J Am Soc Echocardiogr 2015;28:1-39 e14.

15. Liu H, Pozios I, Haileselassie B, et al. Role of global longitudinal strain in predicting outcomes in hypertrophic cardiomyopathy. Am J Cardiol 2017;120:670-5.

16. Bravo PE, Zimmerman SL, Luo HC, et al. Relationship of delayed enhancement by magnetic resonance to myocardial perfusion by positron emission tomography in hypertrophic cardiomyopathy. Circ Cardiovasc Imaging 2013;6:210-7.

17. Yalcin H, Valenta I, Yalcin F, et al. Effect of diffuse subendocardial hypoperfusion on left ventricular cavity size by (13)N-ammonia perfusion PET in patients with hypertrophic cardiomyopathy. Am J Cardiol 2016;118:1908-15.

18. Murphy KP. Machine Learning: A Probabilistic Perspective. Cambridge, MA: MIT Press; 2012.

19. Sokal RR, Rohlf FJ. Introduction to Biostatistics. 2nd ed. New York: Freeman; 1987.

20. Welch BL. The significance of the difference between two means when the population variances are unequal. Biometrika 1938;29:350-62.

21. Pellikka PA, Arruda-Olson A, Chaudhry FA, et al. Guidelines for performance, interpretation, and application of stress echocardiography in ischemic heart disease: from the American Society of Echocardiography. J Am Soc Echocardiogr 2020;33:1-41 e48.

22. Olsson U. Maximum likelihood estimation of the polychoric correlation coefficient. Psychometrika 1979;44:443-60.

23. Bhattacharya M, Lu DY, Kudchadkar SM, et al. Identifying ventricular arrhythmias and their predictors by applying machine learning methods to electronic health records in patients with hypertrophic cardiomyopathy (HCM-VAr-Risk Model). Am J Cardiol 2019;123:1681-9.

24. Schnabel RB, Sullivan LM, Levy D, et al. Development of a risk score for atrial fibrillation (Framingham Heart Study): a community-based cohort study. Lancet 2009;373:739-45.

25. Chamberlain AM, Agarwal SK, Folsom AR, et al. A clinical risk score for atrial fibrillation in a biracial prospective cohort (from the Atherosclerosis Risk in Communities [ARIC] study). Am J Cardiol 2011;107:85-91.

26. Alonso A, Krijthe BP, Aspelund T, et al. Simple risk model predicts incidence of atrial fibrillation in a racially and geographically diverse population: the CHARGE-AF consortium. J Am Heart Assoc 2013;2: e000102.

27. Debonnaire P, Joyce E, Hiemstra Y, et al. Left atrial size and function in hypertrophic cardiomyopathy patients and risk of new-onset atrial fibrillation. Circ Arrhythm Electrophysiol 2017;10:e004052.

28. Guttmann OP, Rahman MS, O'Mahony C, Anastasakis A, Elliott PM. Atrial fibrillation and thromboembolism in patients with hypertrophic cardiomyopathy: systematic review. Heart 2014;100:465-72.

29. Maron BJ, Haas TS, Maron MS, et al. Left atrial remodeling in hypertrophic cardiomyopathy and susceptibility markers for atrial fibrillation identified by cardiovascular magnetic resonance. Am J Cardiol 2014;113: 1394-400.

30. Cochet H, Morlon L, Verge MP, et al. Predictors of future onset of atrial fibrillation in hypertrophic cardiomyopathy. Arch Cardiovasc Dis 2018;111:591-600.

31. Tuluce K, Yakar Tuluce S, Kahya Eren N, et al. Predictors of future atrial fibrillation development in patients with hypertrophic cardiomyopathy: a prospective follow-up study. Echocardiography 2016;33:379-85.

32. Bundy JD, Heckbert SR, Chen LY, Lloyd-Jones DM, Greenland P. Evaluation of risk prediction models of atrial fibrillation (from the Multi-Ethnic Study of Atherosclerosis [MESA]). Am J Cardiol 2020;125:55-62.

33. Harris TB, Launer LJ, Eiriksdottir G, et al. Age, gene/environment susceptibility-Reykjavik Study: multidisciplinary applied phenomics. Am J Epidemiol 2007;165:1076-87.

34. Ikram MA, Brusselle GGO, Murad SD, et al. The Rotterdam Study: 2018 update on objectives, design and main results. Eur J Epidemiol 2017;32:807-50.

35. Al'Aref SJ, Anchouche K, Singh G, et al. Clinical applications of machine learning in cardiovascular disease and its relevance to cardiac imaging. Eur Heart J 2019;40:1975-86.

36. Vallee A, Cinaud A, Blachier V, et al. Coronary heart disease diagnosis by artificial neural networks including aortic pulse wave velocity index and clinical parameters. J Hypertens 2019;37:1682-8.

37. Tsang TS, Gersh BJ, Appleton CP, et al. Left ventricular diastolic dysfunction as a predictor of the first diagnosed nonvalvular atrial fibrillation in 840 elderly men and women. J Am Coll Cardiol 2002;40: 1636-44.

38. Patton KK, Ellinor PT, Heckbert SR, et al. N-terminal pro-B-type natriuretic peptide is a major predictor of the development of atrial fibrillation: the Cardiovascular Health Study. Circulation 2009;120: 1768-74.

39. Schotten U, Neuberger HR, Allessie MA. The role of atrial dilatation in the domestication of atrial fibrillation. Prog Biophys Mol Biol 2003;82: 151-62.

40. Tsang TS, Barnes ME, Gersh BJ, Bailey KR, Seward JB. Risks for atrial fibrillation and congestive heart failure in patients > / = 65 years of age with abnormal left ventricular diastolic relaxation. Am J Cardiol 2004;93: 54-8.

41. Rosenberg MA, Gottdiener JS, Heckbert SR, Mukamal KJ. Echocardiographic diastolic parameters and risk of atrial fibrillation: the Cardiovascular Health Study. Eur Heart J 2012;33:904-12.

42. Tani T, Tanabe K, Ono M, et al. Left atrial volume and the risk of paroxysmal atrial fibrillation in patients with hypertrophic cardiomyopathy. J Am Soc Echocardiogr 2004;17:644-8.

43. Spirito P, Autore C, Formisano F, et al. Risk of sudden death and outcome in patients with hypertrophic cardiomyopathy with benign presentation and without risk factors. Am J Cardiol 2014;113:1550-5.

44. Yang WI, Shim CY, Kim YJ, et al. Left atrial volume index: a predictor of adverse outcome in patients with hypertrophic cardiomyopathy. J Am Soc Echocardiogr 2009;22:1338-43.

45. Tian T, Wang Y, Sun K, et al. Clinical profile and prognostic significance of atrial fibrillation in hypertrophic cardiomyopathy. Cardiology 2013;126:258-64.

46. Losi MA, Betocchi S, Aversa M, et al. Determinants of atrial fibrillation development in patients with hypertrophic cardiomyopathy. Am J Cardiol 2004;94:895-900.

47. Kotecha D, Piccini JP. Atrial fibrillation in heart failure: What should we do? Eur Heart J 2015;36:3250-7.

48. Nagueh SF, Lakkis NM, Middleton KJ, et al. Doppler estimation of left ventricular filling pressures in patients with hypertrophic cardiomyopathy. Circulation 1999;99:254-61.

49. Rosenberg MA, Manning WJ. Diastolic dysfunction and risk of atrial fibrillation: a mechanistic appraisal. Circulation 2012;126:2353-62.

50. Jons C, Joergensen RM, Hassager C, et al. Diastolic dysfunction predicts new-onset atrial fibrillation and cardiovascular events in patients with acute myocardial infarction and depressed left ventricular systolic function: a CARISMA substudy. Eur J Echocardiogr 2010;11:602-7.

51. Pujadas S, Vidal-Perez R, Hidalgo A, et al. Correlation between myocardial fibrosis and the occurrence of atrial fibrillation in hypertrophic cardiomyopathy: a cardiac magnetic resonance imaging study. Eur J Radiol 2010;75:e88-91.

52. Yamaji K, Fujimoto S, Yutani C, et al. Does the progression of myocardial fibrosis lead to atrial fibrillation in patients with hypertrophic cardiomyopathy? Cardiovasc Pathol 2001;10:297-303.

53. Papavassiliu T, Germans T, Fluchter S, et al. CMR findings in patients with hypertrophic cardiomyopathy and atrial fibrillation. J Cardiovasc Magn Reson 2009;11:34.

54. Sivalokanathan S, Zghaib T, Greenland GV, et al. Hypertrophic cardiomyopathy patients with paroxysmal atrial fibrillation have a high burden of left atrial fibrosis by cardiac magnetic resonance imaging. JACC Clin Electrophysiol 2019;5:364-75.

55. Azarbal F, Singh M, Finocchiaro G, et al. Exercise capacity and paroxysmal atrial fibrillation in patients with hypertrophic cardiomyopathy. Heart 2014;100:624-30.

56. Vasquez N, Ostrander BT, Lu DY, et al. Low left atrial strain is associated with adverse outcomes in hypertrophic cardiomyopathy patients. J Am Soc Echocardiogr 2019;32:593-603 e591.

57. Dickinson MG, Lam CS, Rienstra M, et al. Atrial fibrillation modifies the association between pulmonary artery wedge pressure and left ventricular end-diastolic pressure. Eur J Heart Fail 2017;19:1483-90.

58. Luo HC, Dimaano VL, Kembro JM, et al. Exercise heart rates in patients with hypertrophic cardiomyopathy. Am J Cardiol 2015;115:1144-50.

59. Maron BJ, Rowin EJ, Casey SA, et al. Risk stratification and outcome of patients with hypertrophic cardiomyopathy $>= 60$ years of age. Circulation 2013;127:585-93.

60. Bravo PE, Pinheiro A, Higuchi T, et al. PET/CT assessment of symptomatic individuals with obstructive and nonobstructive hypertrophic cardiomyopathy. J Nucl Med 2012;53:407-14.

61. Hurtado-de-Mendoza D, Corona-Villalobos CP, Pozios I, et al. Diffuse interstitial fibrosis assessed by cardiac magnetic resonance is associated with dispersion of ventricular repolarization in patients with hypertrophic cardiomyopathy. J Arrhythm 2017;33:201-7.

62. Kobayashi T, Popovic Z, Bhonsale A, et al. Association between septal strain rate and histopathology in symptomatic hypertrophic cardiomyopathy patients undergoing septal myectomy. Am Heart J 2013;166. 503-11.67.

63. Spirito P, Watson RM, Maron BJ. Relation between extent of left ventricular hypertrophy and occurrence of ventricular tachycardia in hypertrophic cardiomyopathy. Am J Cardiol 1987;60:1137-42.

64. D'Amato R, Tomberli B, Castelli G, et al. Prognostic value of N-terminal pro-brain natriuretic peptide in outpatients with hypertrophic cardiomyopathy. Am J Cardiol 2013;112:1190-6.

65. Gruver EJ, Fatkin D, Dodds GA, et al. Familial hypertrophic cardiomyopathy and atrial fibrillation caused by Arg663His beta-cardiac myosin heavy chain mutation. Am J Cardiol 1999;83. 13H-8H.

66. Ogimoto A, Hamada M, Nakura J, Miki T, Hiwada K. Relation between angiotensin-converting enzyme II genotype and atrial fibrillation in Japanese patients with hypertrophic cardiomyopathy. J Hum Genet 2002;47:184-9.

67. Bongini C, Ferrantini C, Girolami F, et al. Impact of genotype on the occurrence of atrial fibrillation in patients with hypertrophic cardiomyopathy. Am J Cardiol 2016;117:1151-9.

## Supplementary Material

To access the supplementary material accompanying this article, visit *CJC Open* at https://www.cjcopen.ca/ and at https://doi.org/10.1016/j.cjco.2021.01.016.